\pdfoutput=1
\documentclass[11pt]{article}

\usepackage{emnlp2021}

\usepackage{times}
\usepackage{latexsym}

\usepackage[T1]{fontenc}
\usepackage[utf8]{inputenc}

\usepackage{microtype}

\usepackage{amsfonts}
\usepackage{amssymb}
\usepackage{booktabs}
\usepackage{bm}
\usepackage{color}
\usepackage{multirow}
\usepackage[export]{adjustbox}
\usepackage{float}
\usepackage{comment}

\title{Adaptive Information Seeking for Open-Domain Question Answering}

\author{Yunchang Zhu$^{\dagger\S}$, Liang Pang$^{\dagger*}$, Yanyan Lan$^{\diamondsuit*}$, Huawei Shen$^{\dagger\S}$, Xueqi Cheng$^{\ddagger\S}$ \\
    $^{\dagger}$Data Intelligence System Research Center \\ 
    and $^{\ddagger}$CAS Key Lab of Network Data Science and Technology, \\
    Institute of Computing Technology, Chinese Academy of Sciences \\
    $^\S$University of Chinese Academy of Sciences \\
    $^\diamondsuit$Institute for AI Industry Research, Tsinghua University \\
    \texttt{\{zhuyunchang17s, pangliang, shenhuawei, cxq\}@ict.ac.cn} \\
    \texttt{lanyanyan@tsinghua.edu.cn}
}

\begin{document}
\maketitle
\begin{abstract}
Information seeking is an essential step for open-domain question answering to efficiently gather evidence from a large corpus. 
Recently, iterative approaches have been proven to be effective for complex questions, by recursively retrieving new evidence at each step.
However, almost all existing iterative approaches use predefined strategies, either applying the same retrieval function multiple times or fixing the order of different retrieval functions, which cannot fulfill the diverse requirements of various questions. 
In this paper, we propose a novel adaptive information-seeking strategy for open-domain question answering, namely AISO. 
Specifically, the whole retrieval and answer process is modeled as a partially observed Markov decision process, where three types of retrieval operations (e.g., BM25, DPR, and hyperlink) and one answer operation are defined as actions.
According to the learned policy, AISO could adaptively select a proper retrieval action to seek the missing evidence at each step, based on the collected evidence and the reformulated query, or directly output the answer when the evidence set is sufficient for the question.
Experiments on SQuAD Open and HotpotQA fullwiki, which serve as single-hop and multi-hop open-domain QA benchmarks, show that AISO outperforms all baseline methods with predefined strategies in terms of both retrieval and answer evaluations. 
% Moreover, AISO has won the first place on the HotpotQA fullwiki leaderboard since May 10, 2021.
\let\thefootnote\relax\footnotetext{*Corresponding Author}
\end{abstract}

\section{Introduction}
Open-domain question answering (QA) \citep{voorhees1999trec} is a task of answering questions using a large collection of texts (e.g., Wikipedia). 
It relies on a powerful information-seeking method to efficiently retrieve evidence from the given large corpus.

\begin{figure}
    \centering
    \includegraphics[width=0.47\textwidth]{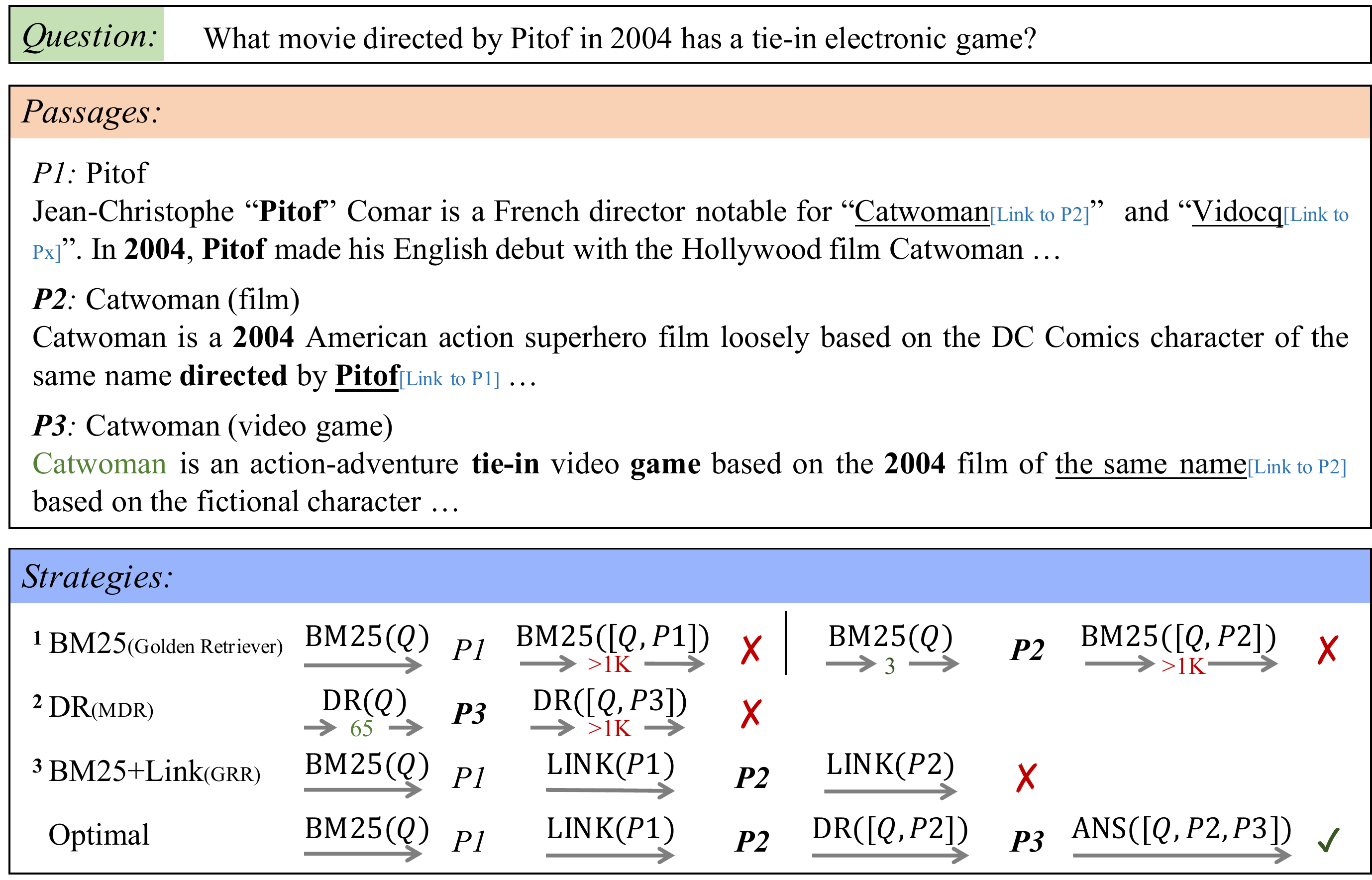}
    \caption{An example derived from HotpotQA development set. P1, P2 and P3 are the most relevant passages, of which P2 and P3 are supporting passages, which are essential to answer the question. Except for the adaptive strategy in the last row, fixed strategy methods such as using BM25 or dense retrieval multiple times and first using BM25 and then entity linking have failed, due to the rank of the remaining supporting passages larger than 1k. The number between two arrows indicates the highest rank of the remaining supporting passages in the retrieval list, unless ranked first.}
    \label{fig:exmaple}
\end{figure}

Traditional open-domain QA approaches mainly follow the two-stage retriever-reader pipeline \citep{chen-etal-2017-reading, yang-etal-2018-hotpotqa, karpukhin-etal-2020-dense}, in which the retriever uses a determinate sparse or dense retrieval function to retrieve evidence, independently from the reading stage. 
But these approaches have limitations in answering complex questions, which need multi-hop or logical reasoning \cite{xiong2021answering}. 

To tackle this issue, iterative approaches have been proposed to recurrently retrieve passages and reformulate the query based on the original question and the previously collected passages. 
Nevertheless, all of these approaches adopt fixed information-seeking strategies in the iterative process.
For example, some works employ a single retrieval function multiple times \citep{das2018multistep, qi-etal-2019-answering, xiong2021answering}, and the other works use a pre-defined sequence of retrieval functions \citep{Asai2020Learning, Dhingra2020Differentiable}.

However, the fixed information-seeking strategies cannot meet the diversified requirements of various problems. 
Taking Figure~\ref{fig:exmaple} as an example, the answer to the question is `Catwoman' in P3.
Due to the lack of essential supporting passages, simply applying BM25/dense retrieval (DR) multiple times (strategy 1 \citep{qi-etal-2019-answering} or 2 \citep{xiong2021answering}), or using the mixed but fixed strategy (strategy 3 \citep{Asai2020Learning}) cannot answer the question.
Specifically, it is hard for \citet{qi-etal-2019-answering} to generate the ideal query `Catwoman game' by considering P1 or P2, thus BM25 \citep{10.1561/1500000019} suffers from the mismatch problem and fails to find the next supporting passage P3.
The representation learning of salient but rare phrases (e.g.~`Pitof') still remains a challenging problem \citep{karpukhin-etal-2020-dense}, which may affect the effectiveness of dense retrieval, i.e.,~the supporting passage P3 is ranked 65, while P1 and P2 do not appear in the top-1000 list at the first step.
Furthermore, link retrieval functions fail when the current passage, e.g.,~P2, has no valid entity links.

Motivated by the above observations, we propose an \textbf{A}daptive \textbf{I}nformation-\textbf{S}eeking approach for \textbf{O}pen-domain QA, namely AISO. 
Firstly, the task of open-domain QA is formulated as a partially observed Markov decision process (POMDP) to reflect the interactive characteristics between the QA model (i.e.,~agent) and the intractable large-scale corpus (i.e.,~environment).
The agent is asked to perform an action according to its state (belief module) and the policy it learned (policy module).
Specifically, the belief module of the agent maintains a set of evidence to form its state.
Moreover, there are two groups of actions for the policy module to choose, 
1) retrieval action that consists of the type of retrieval function and the reformulated query for requesting evidence, 
and 2) answer action that returns a piece of text to answer the question, then completes the process.
Thus, in each step, the agent emits an action to the environment, which returns a passage as the observation back to the agent. 
The agent updates the evidence set and generates the next action, step by step, until the evidence set is sufficient to trigger the answer action to answer the question.
To learn such a strategy, we train the policy in imitation learning by cloning the behavior of an oracle online, which avoids the hassle of designing reward functions and solves the POMDP in the fashion of supervised learning.

Our experimental results show that our approach achieves better retrieval and answering performance than the state-of-the-art approaches on SQuAD Open and HotpotQA fullwiki, which are the representative single-hop and multi-hop datasets for open-domain QA. 
Furthermore, AISO significantly reduces the number of reading steps in the inference stage.
% Finally, detailed analyses on the AISO show the good interpretability of our method.

In summary, our contributions include: 
\begin{itemize}
    \item To the best of our knowledge, we are the first to introduce the adaptive information-seeking strategy to the open-domain QA task;
    \item Modeling adaptive information-seeking as a POMDP, we propose AISO, which learns the policy via imitation learning and has great potential for expansion.
    \item The proposed AISO achieves state-of-the-art performance on two public dataset and wins the first place on the HotpotQA fullwiki leaderboard. Our code is available at \url{https://github.com/zycdev/AISO}.
\end{itemize}

\begin{figure*}
    \centering
    \includegraphics[width=1\textwidth]{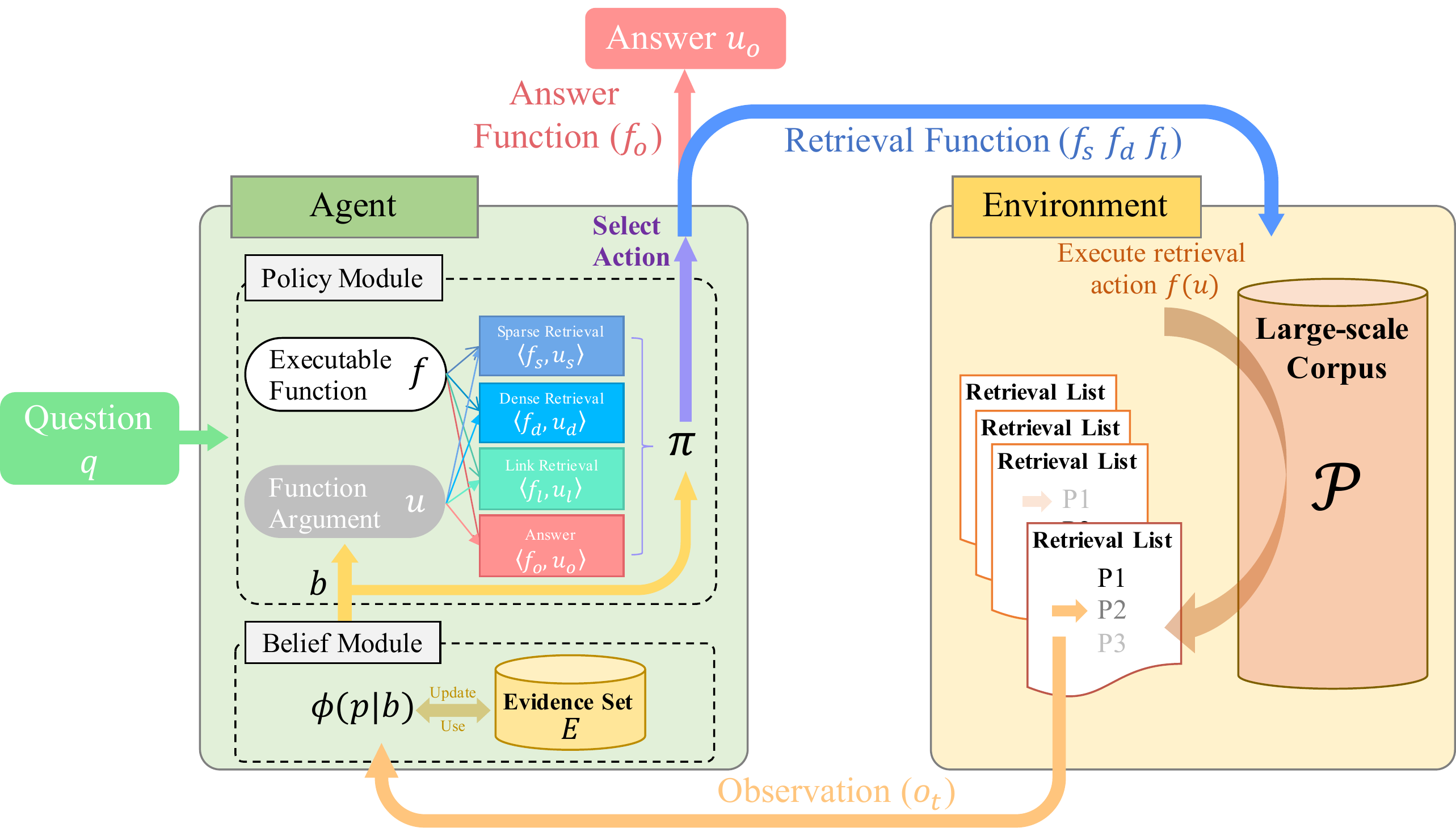}
    \caption{The overview of the AISO.}
    \label{fig:model}
\end{figure*}

\section{Related Work}
Traditional approaches of open-domain QA mainly follow the two-stage retriever-reader pipeline \citep{chen-etal-2017-reading}: a retriever first gathers relevant passages as evidence candidates, then a reader reads the retrieved candidates to form an answer.
In the retrieval stage, most approaches employ a determinate retrieval function and treat each passage independently \citep{wang2018r, lin-etal-2018-denoising, lee-etal-2018-ranking, yang-etal-2018-hotpotqa, pang2019has, lee-etal-2019-latent, pmlr-v119-guu20a, karpukhin-etal-2020-dense, izacard2020leveraging}.
As an extension, some approaches further consider the relations between passages through hyperlinks or entity links and extend evidence with the linked neighbor passages \citep{nie-etal-2019-revealing, das-etal-2019-multi, Zhao2020Transformer-XH}.
However, pipeline approaches retrieve evidence independently from reader, leading to 1) introduce less-relevant evidence to the question, and 2) hard to model the complex question which has high-order relationship between question and evidence.

Instead, recent iterative approaches sequentially retrieve new passages by updating the query inputted to a specific retrieval function at each step, conditioned on the information already gathered.
At each step, \citet{das2018multistep, feldman-el-yaniv-2019-multi, xiong2021answering} reformulate the dense query vector in a latent space, while \citet{ding-etal-2019-cognitive, qi-etal-2019-answering, zhang2020ddrqa, qi2020retrieve} update the natural language query.
After the first step retrieval using TF-IDF, \citet{Asai2020Learning} and \citet{li2020hopretriever} recursively select subsequent supporting passages on top of a hyperlinked passage graph.
Nevertheless, all of these approaches adopt fixed information-seeking strategies, employing the same retrieval function multiple times \citep{das2018multistep, feldman-el-yaniv-2019-multi, xiong2021answering, ding-etal-2019-cognitive, qi-etal-2019-answering, zhang2020ddrqa, qi2020retrieve} or pre-designated sequence of applying retrieval functions \citep{Asai2020Learning, li2020hopretriever}.
Due to the diversity of questions, these fixed strategies established in advance may not be optimal for all questions, or even fail to collect evidence.

\section{Method}
In this section, we first formulate the open-domain QA task as a partially observed Markov decision process (POMDP) and introduce the dynamics of the environment.
Then, we elaborate on how the agent interacts with the environment to seek evidence and answer a question.
Finally, to solve the POMDP, we describe how to train the agent via imitation learning.

\subsection{Open-Domain QA as a POMDP}\label{odqa_pomdp}
Given a question $q$ and a large corpus $\mathcal{P}$ composed of passages, the task of open-domain QA is to collect a set of evidence $E \subset \mathcal{P}$ and answer the question based on the gathered evidence.

The fashion of iterative evidence gathering, proven effective by previous works \citep{das2018multistep, Asai2020Learning, xiong2021answering}, is essentially a sequential decision-making process.

Besides, since the corpus is large, ranging from millions (e.g., Wikipedia) to billions (e.g., the Web), and the input length of a QA model is limited, the QA model can only observe a part of the corpus.
Owing to the above two reasons, we model open-domain QA as a partially observed Markov decision process.

In the POMDP we designed, as shown in Figure~\ref{fig:model}, the agent is the QA model that needs to issue actions to seek evidence from the large-scale corpus hidden in the environment and finally respond to the question.
By executing the received action, the environment can return a retrieved passage to the agent as an observation of the corpus.
Formally, the POMDP is defined by $(\mathcal{S}, \mathcal{A}, \mathcal{O}, \Omega, Z, R)$, where $R$ is the reward function.

% \textbf{Reward.} At each time step $t$, the agent receives a immediate reward $R(s_t, a_t)$, where $R: \mathcal{S} \times \mathcal{A} \mapsto \mathbb{R}$ is a hand-craft reward function and is not in the scope of this paper.

\textbf{Actions:} 
At timestep $t = 0, 1, \cdots, T$, the action $a_t$ in the action space $\mathcal{A} = \mathcal{F} \times \mathcal{U}$ is a request for an executable function $f \in \mathcal{F}$, expressed as $\langle f, u \rangle$, where $u \in \mathcal{U}$ is the text argument that gets passed to $f$.
The space of executable functions $\mathcal{F}$ includes two groups of functions,
1) retrieval function that takes the query $u$ and corpus $\mathcal{P}$ as input and ranks a retrieval list of passages as $\mathcal{P}_{f(u)}$, 
2) answer function that replies to the question $q$ with the answer $u$ and ends the process.
The action $a_t$ is performed following the policy $\Pi$ described in Subsection~\ref{policy}.

\textbf{States:} 
The environment state $s_t$ in the state space $\mathcal{S}$ contains revealing states of retrieval lists of all history retrieval actions.
When the agent issues an action $a_t = \langle f, u \rangle$, $s_t$ will transfer to $s_{t+1}$ governed by a deterministic transition dynamics $\Omega(s_t, a_t)$.
Specifically, $\Omega$ will mark the topmost unrevealed passage in the retrieval list $\mathcal{P}_{f(u)}$ as revealed. 
If the environment has never executed $a_t$ before, it will first search and cache $\mathcal{P}_{f(u)}$ for possible repeated retrieval actions in the future.

\textbf{Observations:}
On reaching the new environment state $s_{t+1}$, the environment will return an observation $o_{t+1}$ from the observation space $\mathcal{O} = \{q\} \cup \mathcal{P}$, governed by the deterministic observation dynamics $Z$. 
At the initial timestep, the question $q$ will returned as $o_0$.
In other cases, $Z$ is designed to return only the last passage marked as revealed in $\mathcal{P}_{f(u)}$ at a time.
For example, if the action $\langle f, u \rangle$ is received for the $k$th time, the $k$th passage in $\mathcal{P}_{f(u)}$ will be returned.

\subsection{Agent}\label{agent}
The agent interacts with the environment to collect evidence for answering the question.
Without access to the environment state $s_t$, the agent can only perform sub-optimal actions based on current observations. 
It needs to build its belief $b_t$ in the state that the environment may be in, based on its experience $h_t = (o_0, a_0, o_1, \cdots, a_{t-1}, o_t)$.
Therefore, the agent consists of two modules: \textbf{belief module} $\Phi$ that generates the belief state $b_t = \Phi(h_t)$ from the experience $h_t$, and \textbf{policy module} $\Pi$ that prescribes the action $a_t = \Pi(b_t)$ to take for current belief state $b_t$.

Both belief and policy modules are constructed based on pretrained Transformer encoders \citep{Clark2020ELECTRA}, respectively denoted as $\Psi^{belief}$ and $\Psi^{policy}$, which encode each inputted token into a $d$-dimensional contextual representation.
The input of both encoders is a belief state, formatted as ``\verb|[CLS]| \verb|[YES]| \verb|[NO]| \verb|[NONE]| $\mathrm{question}$ \verb|[SEP]| $\mathrm{title}_{o}$ \verb|[SOP]| $\mathrm{content}_{o}$ \verb|[SEP]| $\mathrm{title}_1$ \verb|[SOP]| $\cdots$ $\mathrm{content}_{|E|}$ \verb|[SEP]|'', where the subscript $_o$ denotes the observation passage, and the others passages come from the collected evidence set $E$, \verb|[SOP]| is a special token to separate the title and content of a passage, \verb|[YES]| and \verb|[NO]| are used to indicate yes/no answer, and \verb|[NONE]| is generally used to indicate that there is no desired answer/query/evidence.
In this way, the self-attention mechanism across the concatenated sequence allows each passage in the input to interact with others, which has been shown crucial for multi-hop reasoning \citep{wang-etal-2019-multi-hop}.

\subsubsection{Belief Module}\label{belief}
The belief module $\Phi$ transforms the agent's experience $h_t$ into a belief state $b_t$ by maintaining a set of evidence $E_{t-1}$.
At the end of the process, the evidence set $E$ is expected to contain sufficient evidence necessary to answer the question and no irrelevant passage.
In the iterative process, the agent believes that all the passages in $E$ may help answer the question.
In other words, those passages that were observed but excluded from the evidence set, i.e., $o_{1:t-1} \setminus E_{t-1}$, are believed to be irrelevant to the question.

For simplicity, assuming that the negative passages $o_{1:t-1} \setminus E_{t-1}$ and action history $a_{<t}$ are not helpful for subsequent decision-making, the experience $h_t$ is equivalent to $\{q, o_t\} \cup E_{t-1}$.
Thus, let $C_t = E_{t-1} \cup \{o_t\}$ be the current candidate evidence set, then the original question and current evidence candidates can form the belief state $b_t$ as
\begin{equation}
    b_t = \Pi(h_t) = \langle q, C_t \rangle = \langle q, E_{t-1} \cup \{o_t\} \rangle.
\end{equation}
At the beginning, the belief state $b_0$ is initialized to $\langle q, \varnothing \rangle$, and the evidence set $E_0$ is initialized to $\varnothing$.

To maintain the essential evidence set $E_t$, we use a trainable scoring function $\phi(p|b_t)$ to identify each evidence candidate $p \in C_t$.
Specifically, each passage is represented as the contextual representation of the special token \verb|[SOP]| in it, which is encoded by $\Psi^{belief}$.
Then, the representation of each candidate is projected into a score through a linear layer.
Besides, we use a pseudo passage $p_0$, represented as \verb|[None]|, to indicate the dynamic threshold of the evidence set.
In this way, after step $t$, the evidence set is updated as
\begin{equation}
    E_t = \{p_i | \phi(p_i|b_t) > \phi(p_0|b_t), p_i \in C_t\}.
\end{equation}
It is worth noting that these evidence candidates are scored jointly since encoded together in the same input, different from conventional rerankers that score separately.

\subsubsection{Policy Module}\label{policy}
The policy module $\Pi$ decides the next action $a_{t}$ to be taken based on the current belief state $b_t$.
In this paper, we equipped the agent with three retrieval functions and one answer function, which means that the action space $\mathcal{A}$ consists of three types of retrieval actions and one type of answer actions.
However, unlike the finite space of executable functions $\mathcal{F}$, the space of function arguments $\mathcal{U}$ includes all possible natural-language queries and answers.
To narrow the search space, for each executable function, we employ a suggester to propose a plausible query or answer as the argument passed to the function.
Finally, we apply an action scoring function in the narrowed action space and select the action with the highest score.

\paragraph{Equipped Functions}
Formally, the space of executable functions is defined as $\mathcal{F} = \{f_s, f_d, f_l, f_o\}$.

Among them, except $f_o$ is the answer function used to reply to the question, the rest are three distinct off-the-shelf retrieval functions (RF) used to explore the corpus.
$f_s$ is a sparse RF, implemented as BM25 \citep{10.1561/1500000019}. 
It performs well when the query is concise and contains highly selective keywords but often fails to capture the semantics of the query.
$f_d$ is a dense RF, implemented as MDR \citep{xiong2021answering} for multi-hop questions, and DPR \citep{karpukhin-etal-2020-dense} for single-hop questions.
Dense RFs can capture lexical variations and semantic relationships, but they struggle when encountering out-of-vocabulary words. 
$f_l$ is a link RF, implemented as hyperlink.
When hyperlink markups are available in a source passage, it can readily map a query (i.e., anchor text) to the target passage.

\paragraph{Argument Generation}
The space of function arguments $\mathcal{U}$, composed of textual queries and answers, is too large to perform an exhaustive search due to the complexity of natural language.
To reduce the search complexity, inspired by \citet{yao-etal-2020-keep}, we employ four argument generators to generate the most plausible query/answer for the equipped functions.

$g_o$ is a trainable reading comprehension model for $f_o$.
It is a span extractor built upon the contextual representations outputted by the encoder $\Psi^{policy}$.
Like conventional extractive reading comprehension models \citep{yang-etal-2018-hotpotqa, Clark2020ELECTRA}, $g_o$ uses the contextual representations to calculate the start and end positions of the most plausible answer $u_o$.
If the current context $C_t$ is insufficient to answer the question, the special token \verb|[NONE]| will be extreacted.

$g_s$ is a query reformulation model for $f_s$.
In this work, we directly employ the well-trained query reformulator from \citet{qi-etal-2019-answering} for multi-hop questions, which takes the belief state $b_t$ as input and outputs a span of the input sequence as the sparse query $u_s$.
As for single-hop questions, since there exists no off-the-shelf multi-step query reformulator, we leave $g_s$ as an identity function that returns the original question directly.
In this case, requesting the same RF multiple times is equivalent to traverse the retrieval list of original question.

$g_d$ is a query reformulator for $f_d$.
For multi-hop questions, $g_d$ concatenates the question $q$ and the passage with the highest score in evidence set $E_t$ as the dense query $u_d$, the same as the input of MDR \citep{xiong2021answering}.
If $E_t$ is empty, $u_d$ is equal to the question $q$.
Similar to $g_s$, $g_d$ for single-hop questions also leaves original questions unchanged.

$g_l$ is a trainable multi-class classifier for $f_l$.
It selects the most promising anchor text from the belief state $b_t$.
To enable rejecting all anchors, \verb|[NONE]| is also treated as a candidate anchor.
$g_l$ shares the encoder $\Psi^{policy}$, where each anchor is represented by the average of contextual representations of its tokens.
Upon $\Psi^{policy}$, we use a linear layer to project the hidden representations of candidate anchors to real values and select the anchor with the highest value as the link query $u_l$.

In this way, the action space is narrowed down to $\check{A} = \{\langle f_s, u_s \rangle, \langle f_d, u_d \rangle, \langle f_l, u_l \rangle, \langle f_o, u_o \rangle\}$.

\paragraph{Action Selection}
The action scoring function $\pi$ is also built upon the output of $\Psi^{policy}$.
To score an action $\langle f, u \rangle$ for current belief state $b_t$, an additional two-layer ($3d\times4d\times1$) MLP, with a ReLU activation in between, projects the concatenated representation of $b_t$, executable function $f$, and function argument $u$, i.e., $\bm{v}_\mathrm{[CLS]}$, $\bm{w}_f$, and $\bm{v}_u$, into a real value.
$\bm{w}_f \in \mathbb{R}^d$ is a trainable embedding for each executable function, the same dimension as the token embedding.
$\bm{v}_u$ is specific for each function.
Since $u_s$, $u_l$ and $u_o$ have explicit text span in the $b_t$, thus their $\bm{v}_{u}$ are the averages of their token representations.
As for $u_d$, if $g_d$ does not expand the original question, $\bm{v}_{u_d}$ is the contextual representation of \verb|[NONE]|. 
Otherwise, $\bm{v}_{u_d}$ is the \verb|[SOP]| of the passage concatenated to the question.

In short, the next action is selected from the narrowed action space $\check{A}$ by the scoring function $\pi$,
\begin{equation}
    a_t = \Pi(b_{t}) = \mathop{\arg\max}_{a \in \check{A}} \pi(a|b_{t}).
\end{equation}

\subsection{Training}\label{training}
In the agent, in addition to the encoders $\Psi^{belief}$ and $\Psi^{policy}$, we need to train the evidence scoring function $\phi$, link classifier $g_l$,  answer extractor $g_o$, and action scoring function $\pi$, whose losses are $L_\phi$, $L_l$, $L_o$, and $L_\pi$.
Since the policy module is dependent on the belief module, we train the agent jointly using the following loss function,
\begin{equation}
    L = L_\phi + L_l + L_o + L_\pi.
\end{equation}

Unlike $\phi$, $g_l$ and $g_o$ that can be trained in supervised learning through human annotations in QA datasets, the supervision signal for $\pi$ is hard to be derived directly from QA datasets.
Even though policies are usually trained via reinforcement learning, reinforcement learning algorithms \citep{sutton2000policy, mnih2015human} are often sensitive to the quality of reward functions. 
For a complex task, the reward function $R$ is often hard to specify and exhaustive to tune.
Inspired by \citet{choudhury2017adaptive}, we explore the use of imitation learning (IL) by querying a model-based oracle online and imitating the action $a^{\star}$ chose by the oracle, which avoids the hassle of designing $R$ and solves the POMDP in the fashion of supervised learning.
Thus, the loss of $\pi$ is defined as the cross entropy,
\begin{equation}
    L_\pi = -\log \frac{e^{\pi(a^{\star}|b)}}{\sum_{a \in \check{A}} e^{\pi(a|b)}},
\end{equation}
where $b$ is the belief state of the agent.

The link classifier $g_l$ and the answer extractor $g_o$ are also optimized with multi-class cross-entropy losses.
For $g_l$, denoting its loss as $L_l$, the classification label is set to the anchor text that links to a gold supporting passage, if there is no such anchor, then the pseudo hyperlink \verb|[NONE]| is labeled.
$g_o$ is trained as a classifier of start and end position following previous work \citep{Clark2020ELECTRA}, denoting its loss as $L_o$.
Considering the belief state $b = \langle q, \{p_1, p_2, \cdots, p_{|C|}\} \rangle$, the ListMLE \citep{10.1145/1390156.1390306} ranking loss of the evidence scoring function $\phi$ is defined as the negative log likelihood of the ground truth permutation,
\begin{equation}
  L_{\phi}(\bm{y}, b) = -\log P(\tau_{\bm{y}}|\{\phi(p_i|b)\}_{i=0}^{|C|}),
\end{equation}
where $\bm{y}$ is the relevance label of $\{p_0, p_1, \cdots, p_{|C|}\}$ and $\tau_{\bm{y}}$ is their ground truth permutation.
To learn the dynamic threshold $\phi(p_0|b)$, we set the relevance label of the pseudo passage $p_0$ to $\bm{y}_0 = 0.5$. And passages in $C$ are labeled as 1/0 according to whether they are gold supporting passages.

\paragraph{Model-based Oracle}
The model-based oracle has full access to the environment and can foresee the gold evidence and answer of every question, which means that the oracle can infer the rank of a supporting passage in the retrieval list of any retrieval action.
Thus, given a state, the oracle can easily select a near-optimal one from candidate actions according to a greedy policy $\pi^{\star}$.
Specifically, if all gold evidence is collected and the argument of an answer action is a correct answer, the oracle will select the answer action.
Otherwise, the oracle will use a greedy algorithm to select the retrieval action that helps to gather a missing passage of evidence in the fewest steps.

\paragraph{Belief States Sampling}
We train the agent on sampled belief states instead of long trajectories.
In every epoch, one belief state is sampled for each question.
To sample a belief state $\langle q, C \rangle$, we first uniformly sample a subset from $q$'s gold evidence as $C$, which could be an empty set.
However, at testing time, it is impossible for the candidate evidence set $C$ to contain only gold evidence.
To alleviate the mismatch of the state distribution between training and testing, we inject a few negative passages into $C$ and shuffle them.
We treat the first passage in the candidate set as the observation, and the others as evidence collected before.

The distribution of injected negative passages can affect the test performance.
In this work, to make it simple, we sample 0\textasciitilde2 passages from all top-ranked negative passages in retrieval lists of $f_s$, $f_d$, and $f_l$.

\section{Experiments}
We evaluate AISO and baselines on two Wikipedia-sourced benchmarks. 
We first introduce the experimental setups, then describe the experimental results on evidence gathering and question answering. 
Furthermore, detailed analyses are discussed. 

\subsection{Experimental Setup}

\paragraph{Data}
HotpotQA \citep{yang-etal-2018-hotpotqa}, a multi-hop QA benchmark. We focus on its fullwiki (open-domain) setting\footnote{https://hotpotqa.github.io/wiki-readme.html}.
It requires gathering two supporting passages (paragraphs) to answering a question, given the introductory (first) paragraphs of 5M Wikipedia articles dumped on October 1, 2017. 

SQuAD Open \citep{chen-etal-2017-reading}, a single-hop QA benchmark, whose questions are from the SQuAD dataset \citep{rajpurkar-etal-2016-squad} and can be answered based on a single passage. 
We preprocess the Wikipedia dump on December 21, 2016 and extract hyperlinks using WikiExtractor\footnote{https://github.com/attardi/wikiextractor. We do not use the processed data provided by \citet{chen-etal-2017-reading} because it removed the hyperlinks required by our link RF.}.
Following \citet{karpukhin-etal-2020-dense}, we split articles into some disjoint passages, resulting in 20M passages in total. 
We add two extra hyperlinks to each passage, one linking to its previous passage in the article, the other to the next passage.

\paragraph{Metrics}
To test whether the top-2 passages in the evidence set exactly cover both gold supporting passages, we use Supporting Passage Exact Match (P EM) as the evaluation metric following \citep{Asai2020Learning}. 
To test the performance of answer extraction, we use EM and F1 as our metrics following \citep{yang-etal-2018-hotpotqa}.

\paragraph{Implementation Details}
For sparse retrieval, we index all passages in the corpus with Elasticsearch and implement BM25 following \citet{qi-etal-2019-answering}\footnote{https://github.com/qipeng/golden-retriever}.
For dense retrieval, we leverage the trained passage encoder and query encoder from \citet{karpukhin-etal-2020-dense}\footnote{https://github.com/facebookresearch/DPR, the multi-set version is used} and \citet{xiong2021answering}\footnote{https://github.com/facebookresearch/multihop\_dense \_retrieval} and index all passage vectors using FAISS \citep{johnson2019billion} offline.
During training, we use the HNSW-based index for efficient low-latency retrieval; in test time, we use the exact inner product search index for better retrieval results.
For link retrieval, the filtered hyperlinks are used, whose targets have to be another article from this dump.

Based on Huggingface Transformers \citep{wolf-etal-2020-transformers}, we use ELECTRA \citep{Clark2020ELECTRA} ($d=768/1024$ for base/large)\footnote{Many recent approaches are based on ELECTRA, so we use ELECTRA for fair comparison.} as the initializations for our encoders $\Psi^{belief}$ and $\Psi^{policy}$.
The maximum number of passages inputted into the encoders is set to 3 and the length of input tokens is limited to 512.
To avoid the high confidence passages from being truncated, we input the passages of evidence in descending order of their belief scores from the previous step.

To accelerate the model training, for the first 24 epochs, $\Psi^{belief}$ and $\Psi^{policy}$ share parameters, for the next 6 epochs, they are trained separately. 
The batch size is 32.
We use Adam optimization with learning rate $2 \times 10^{-5}$.
To select the best agent (QA model), we first save several checkpoints that perform well on heuristic single-step metrics, such as action accuracy.
Then we choose the one that performs best in the whole process on the development set.
In test time, the number of interaction steps is limited to $T$. 
We set the maximum number of steps to $T=1000$ if not specified.
Once the agent has exhausted its step budget, it is forced to answer the question.

\subsection{Results}

\begin{table}
\centering
\scalebox{0.7}{
\begin{tabular}{llcc} 
\toprule
Strategy                                 & Method                                      & P EM                      & \# read        \\ 
\midrule
\multirow{2}{*}{$f_s$}                   & BM25                                        & 11.11                     & 2                   \\
                                         & BM25 + Reranker                           & 29.60                     & 20                  \\ 
\hline
$f_d$                                    & DPR \citep{karpukhin-etal-2020-dense}        & 14.18                     & 2                   \\ 
\hline
%\multirow{2}{*}{-}                       & AISO$_\mathrm{base}^{\mathrm{belief}}$$^{^{\dagger}}$  & 74.62                     & \multirow{2}{*}{-}  \\
%                                         & AISO$_\mathrm{base}^{\mathrm{belief}}$$^{^{\dagger}}$ & 77.91                     &                     \\ 
\multirow{2}{*}{$f_s \circ f_l$}         & Semantic Retrieval$^{\ast\diamondsuit}$       & 69.35                     & 39.4                \\
                                         & Entity Centric IR$^{\ast\heartsuit}$       & 34.90                     & -                   \\ 
\hline
\hline
$f_s \circ f_s$                          & GoldEn Retriever$^\clubsuit$                          & 47.77                     & 10                  \\ 
\hline
\multirow{3}{*}{$f_d \circ f_d$}         & MDR \citep{xiong2021answering}                                        & 64.52                     & 2                   \\
                                         & MDR + Reanker$^{\dagger\ast}$                     & 81.20                     & $\ge$200            \\
                                         & Ballen$^{\dagger\ast}$ \citep{khattab2021baleen}                             & 86.70                     & -                   \\ 
\hline
\multirow{3}{*}{$f_s^n$}                 & CogQA$^{\ast}$ \citep{ding-etal-2019-cognitive}                             & 57.80                     & -                   \\
                                         & DDRQA$^{\dagger\ast}$ \citep{chen-etal-2017-reading}                          & 79.80                     & -                   \\
                                         & IRRR$^{\dagger\ast}$ \citep{qi2020retrieve}                      & 84.10                     & $\ge$150            \\ 
\hline
\multirow{4}{*}{$f_s \circ f_l^{n-1}$}   & GRR$^{\dagger\ast}$ \citep{Asai2020Learning}                       & 75.70                     & $\ge$500            \\
                                         & HopRetriever$^{\dagger\ast}$ \citep{li2020hopretriever}              & 82.54                     & $\ge$500            \\
                                         & HopRetriever-plus$^{\dagger\ast}$   & 86.94                     & $>$500              \\
                                         & TPRR$^{\dagger\ast}$ \citep{TPRR}                      & \multicolumn{1}{l}{86.19} & $\ge$500            \\ 
\hline
$(f_s \parallel f_d)^n$                  & DrKit$^{\ast}$ \citep{Dhingra2020Differentiable}                             & 38.30                     & -                   \\ 
\hline
% $(f_s^2 \parallel f_d^2) \circ f_l$      & Ensemble + Reranker$^{\dagger}$                                          & 77.91                     & 68.9                \\ 
% \hline
\multirow{2}{*}{$(f_s|f_d|f_l)_{\Pi}^n$} & AISO$_\mathrm{base}$                        & 85.69                     & 36.7                \\
                                         & AISO$_\mathrm{large}$                       & \textbf{88.17}            & 35.7                \\
\bottomrule
\end{tabular}}
\caption{Evidence gathering performance and reading cost on the HotpotQA fullwiki development set. 
% $\dagger$ independently reranks all passage candidates returned by retrieval functions.
% $\ddagger$ reranks all potential evidence sets.
% $\wr$ maintains a graph of evidence and reranks each passage in it, conditioned on its neighbors.
% The method without these marks does not employ additional reranker model.
The symbol $\dagger$ denotes the baseline methods use the large version of pretrained language models comparable to our AISO$_\mathrm{large}$.
The results with $\ast$ are from published papers, otherwise they are our implementations.
The symbol $\circ$ denotes sequential apply RFs, $f^n$ denotes apply the RF $f$ multiple times, $||$ denotes combining the results of different RFs, and $(\cdot|\cdot)_\Pi$ means choosing one of RFs to use according to the policy $\Pi$.
$\diamondsuit$: \citep{nie-etal-2019-revealing},
$\heartsuit$: \citep{qi-etal-2019-answering},
$\clubsuit$: \citep{qi-etal-2019-answering}
}
\label{tab:hotpotqa_ret}
\end{table}

\begin{table*}
\centering
\scalebox{0.8}{\begin{tabular}{lcccccccccccc} 
\toprule
\multirow{3}{*}{Method} & \multicolumn{6}{c}{Dev}                                                                       & \multicolumn{6}{c}{Test}                                                                                                                             \\ 
\cmidrule(r){2-7}\cmidrule(r){8-13}
                        & \multicolumn{2}{c}{Ans}       & \multicolumn{2}{c}{Sup}       & \multicolumn{2}{c}{Joint}     & \multicolumn{2}{c}{Ans}                                  & \multicolumn{2}{c}{Sup}                & \multicolumn{2}{c}{Joint}                        \\ 
\cmidrule(r){2-3}\cmidrule(r){4-5}\cmidrule(r){6-7}\cmidrule(r){8-9}\cmidrule(r){10-11}\cmidrule(r){12-13}
                        & EM            & F1            & EM            & F1            & EM            & F1            & EM                                       & F1            & EM                     & F1            & EM                     & F1                      \\ 
\midrule
Semantic Retrieval \citep{nie-etal-2019-revealing}      & 46.5          & 58.8          & 39.9          & 71.5          & 26.6          & 49.2          & 45.3                                     & 57.3          & 38.7                   & 70.8          & 25.1                   & 47.6                    \\
GoldEn Retriever \citep{qi-etal-2019-answering}       & -             & -             & -             & -             & -             & -             & 37.9                                     & 49.8          & 30.7                   & 64.6          & 18.0                   & 39.1                    \\
CogQA \citep{ding-etal-2019-cognitive}                  & 37.6          & 49.4          & 23.1          & 58.5          & 12.2          & 35.3          & 37.1                                     & 48.9          & 22.8                   & 57.7          & 12.4                   & 34.9                    \\
DDRQA$^\dagger$ \citep{zhang2020ddrqa}                  & 62.9          & 76.9          & 51.3          & 79.1          & -             & -             & 62.5                                     & 75.9          & 51.0                   & 78.9          & 36.0                   & 63.9                    \\
IRRR+$^{\dagger\ast}$ \citep{qi2020retrieve}                  & -             & -             & -             & -             & -             & -             & 66.3                                     & 79.9          & 57.2                   & 82.6          & 43.1                   & 69.8                    \\
MUPPET \citep{feldman-el-yaniv-2019-multi}                 & 31.1          & 40.4          & 17.0          & 47.7          & 11.8          & 27.6          & 30.6                                     & 40.3          & 16.7                   & 47.3          & 10.9                   & 27.0                    \\
MDR$^\dagger$ \citep{xiong2021answering}                    & 62.3          & 75.1          & 56.5          & 79.4          & 42.1          & 66.3          &  62.3 & 75.3          & 57.5                   & 80.9          & 41.8                   & 66.6                    \\
GRR$^\dagger$ \citep{Asai2020Learning}                    & 60.5          & 73.3          & 49.2          & 76.1          & 35.8          & 61.4          & 60.0                                     & 73.0          & 49.1                   & 76.4          & 35.4                   & 61.2                    \\
HopRetriever$^\dagger$ \citep{li2020hopretriever}           & 62.2          & 75.2          & 52.5          & 78.9          & 37.8          & 64.5          & 60.8                                     & 73.9          & 53.1                   & 79.3          & 38.0                   & 63.9                    \\
HopRetriever-plus$^\dagger$ \citep{li2020hopretriever}      & 66.6          & 79.2          & 56.0          & 81.8          & 42.0          & 69.0          & 64.8                                     & 77.8          & 56.1                   & 81.8          & 41.0                   & 67.8                    \\
EBS-Large$^\ast$       & -             & -             & -             & -             & -             & -             & 66.2                                     & 79.3          & 57.3                   & 84.0          & 42.0                   & 70.0                    \\
TPRR$^{\dagger\ast}$ \citep{TPRR}          & 67.3          & 80.1          & 60.2          & 84.5          & 45.3          & 71.4          & 67.0                                     & 79.5          & 59.4                   & 84.3          & 44.4                   & 70.8                    \\ 
\hline
AISO$_\mathrm{base}$    & 63.5          & 76.5          & 55.1          & 81.9          & 40.2          & 66.9          & -                                        & -             & -                      & -             & -                      & -                       \\
AISO$_\mathrm{large}$   & \textbf{68.1} & \textbf{80.9} & \textbf{61.5} & \textbf{86.5} & \textbf{45.9} & \textbf{72.5} & \textbf{67.5}                            & \textbf{80.5} & \textbf{\textbf{61.2}} & \textbf{86.0} & \textbf{\textbf{44.9}} & \textbf{\textbf{72.0}}  \\
\bottomrule
\end{tabular}}
\caption{Answer extraction and supporting sentence identification performance on HotpotQA fullwiki. 
The methods with $\dagger$ use the large version of pretrained language models comparable to AISO$_\mathrm{large}$.
The results marked with $\ast$ are from the official leaderboard otherwise originated from published papers.
}
\label{tab:hotpotqa_ans}
\end{table*}

\begin{table}
\centering
\scalebox{0.75}{
\begin{tabular}{lccc} 
\toprule
Method                                                  & EM            & F1            & \# read  \\ 
\midrule
DrQA \citep{chen-etal-2017-reading}                     & 27.1          & -             & 5        \\
Multi-passage BERT \citep{wang-etal-2019-multi-passage} & 53.0          & 60.9          & 100      \\
DPR \citep{karpukhin-etal-2020-dense}                   & 29.8          & -             & 100      \\
BM25+DPR \citep{karpukhin-etal-2020-dense}              & 36.7          & -             & 100      \\
Multi-step Reasoner \citep{das2018multistep}            & 31.9          & 39.2          & 5        \\
MUPPET \citep{feldman-el-yaniv-2019-multi}              & 39.3          & 46.2          & 45       \\
GRR$^\dagger$ \cite{Asai2020Learning}                   & 56.5          & 63.8          & $\ge500$ \\
SPARTA$^\dagger$ \citep{zhao2020sparta}                 & 59.3          & 66.5          & -        \\
IRRR$^\dagger$ \citep{qi2020retrieve}                   & 56.8          & 63.2          & $\ge150$ \\
\hline
AISO$_\mathrm{large}$                                   & \textbf{59.5} & \textbf{67.6} & 24.8 \\
\bottomrule
\end{tabular}}
\caption{Question answering performance on SQuAD Open benchmark. $\dagger$ denotes the methods use the large pretrained language models comparable to AISO$_\mathrm{large}$.}
\label{tab:squad_ans}
\end{table}

\paragraph{Evidence Gathering}
We first evaluate the performance and reading cost on the evidence gathering, illustrating the effectiveness and efficiency of AISO. 
In Table~\ref{tab:hotpotqa_ret}, we split evidence gathering methods into different groups according to their strategies. 
Moreover, the first three groups are the traditional pipeline approaches, and the others are iterative approaches.

For effectiveness, we can conclude that 1) almost all the iterative approaches perform better than the pipeline methods, 2) the proposed adaptive information-seeking approach AISO$_\mathrm{large}$ outperforms all previous methods and achieves the state-of-the-art performance.
Moreover, our AISO$_\mathrm{base}$ model outperforms some baselines that use the large version of pretrained language models, such as HopRetriever, GRR, IRRR, DDRQA, and MDR.

For efficiency, the cost of answering an open-domain question includes the retrieval cost and reading cost.
Since the cost of reading a passage along with the question online is much greater than the cost of a search, the total cost is linear in \#~read, reported in the last column of Table~\ref{tab:hotpotqa_ret}.
\#~read means the total number of passages read along with the question throughout the process, which is equal to the adaptive number of steps.
We can find that the number of read passages in AISO model, i.e., the  is about 35, which is extremely small than the competitive baselines (P EM > 80) that need to read at least 150 passages. 
That is to say, our AISO model is efficient in practice.

\paragraph{Question Answering}
Benefit from high-performance evidence gathering, as shown in Tables \ref{tab:hotpotqa_ans} and \ref{tab:squad_ans}, AISO outperforms all existing methods across the evaluation metrics on the HotpotQA fullwiki and SQuAD Open benchmarks.
This demonstrates that AISO is applicable to both multi-hop questions and single-hop questions.
Notably, on the HotpotQA fullwiki blind test set\footnote{https://hotpotqa.github.io. As of September 2021, AISO is still at the top of the fullwiki leaderboard.},  AISO$_\mathrm{large}$ significantly outperforms the second place TPRR \citep{TPRR} by 2.02\% in Sup F1 (supporting sentence identification) and 1.69\% on Joint F1.
% Besides, we can see that AISO$_\mathrm{base}$ outperforms some methods (DDRQA, MDR, GRR and HopRetriever) with large pretrained model on supporting fact identification.

\subsection{Analysis}

\begin{table}
\centering
\scalebox{0.8}{
\begin{tabular}{lccc} 
\toprule
Model                             & P EM  & Ans F1 & \# read  \\ 
\midrule
AISO$_\mathrm{base}$              & 85.69 & 76.45  & 36.64    \\ 
\hline
%w/o deletion                      & 85.01 & 76.14  & 36.21    \\
%w.~$\pi_{\mathrm{rand}}$          & 77.76 & 73.15  & 20.76    \\
w.~$\phi^{\star}$                 & 97.52 & 79.99  & 40.01    \\
w.~$\phi^{\star}$~+~$\pi^{\star}$ & 98.88 & 80.34  & 8.92     \\ 
\hline
$f_s^t$                           & 68.51 & 67.33  & 58.74    \\
$f_d^t$                           & 79.80 & 72.91  & 68.63    \\
% w/o $f_s$
$(f_d|f_l)_{\Pi}^{n}$             & 83.97 & 74.93  & 61.41    \\
% w/o $f_d$
$(f_s|f_l)_{\Pi}^{n}$             & 82.44 & 74.44  & 37.76    \\
% w/o $f_l$  
$(f_s|f_d)_{\Pi}^{n}$             & 79.66 & 73.36  & 42.01    \\
\bottomrule
\end{tabular}}
\caption{Analysis experiments on HotpotQA fullwiki.}
\label{tab:ablation}
\end{table}

We conduct detailed analysis of AISO$_\mathrm{base}$ on the HotpotQA fullwiki development set. 

\paragraph{The effect of the belief and policy module}
As shown in the second part of Table~\ref{tab:ablation}, we examine the variations of AISO with the oracle evidence scoring function $\phi^{\star}$ or oracle action scoring function $\pi^{\star}$, which are key components of the belief and policy module.
When we replace our learned evidence scoring function with $\phi^{\star}$ that can identify supporting passage perfectly, the performance increase a lot while the reading cost do not change much.
This means that the belief module has a more impact on the performance than the cost. 
If we further replace the learned $\pi$ with $\pi^{\star}$, the cost decreases a lot.
This shows that a good policy can greatly improve the efficiency.

\paragraph{The impact of retrieval functions}
As shown in the last part Table~\ref{tab:ablation}, the use of a single RF, such as $f_s^t$ and $f_d^t$, leads to poor performance and low efficiency.
Moreover, lack of any RF will degrade performance, which illustrates that all RFs contribute to performance.
Specifically, although the link RF $f_l$ cannot be used alone, it contributes the most to performance and efficiency.
Besides, the sparse RF $f_s$ may be better at shortening the information-seeking process than the dense RF $f_d$, since removing $f_s$ from the action space leads to the number of read passages increase from 36.64 to 61.41. 
We conjecture this is because $f_s$ can rank the evidence that matches the salient query very high.

\begin{figure}
    \centering
    \includegraphics[width=0.48\textwidth]{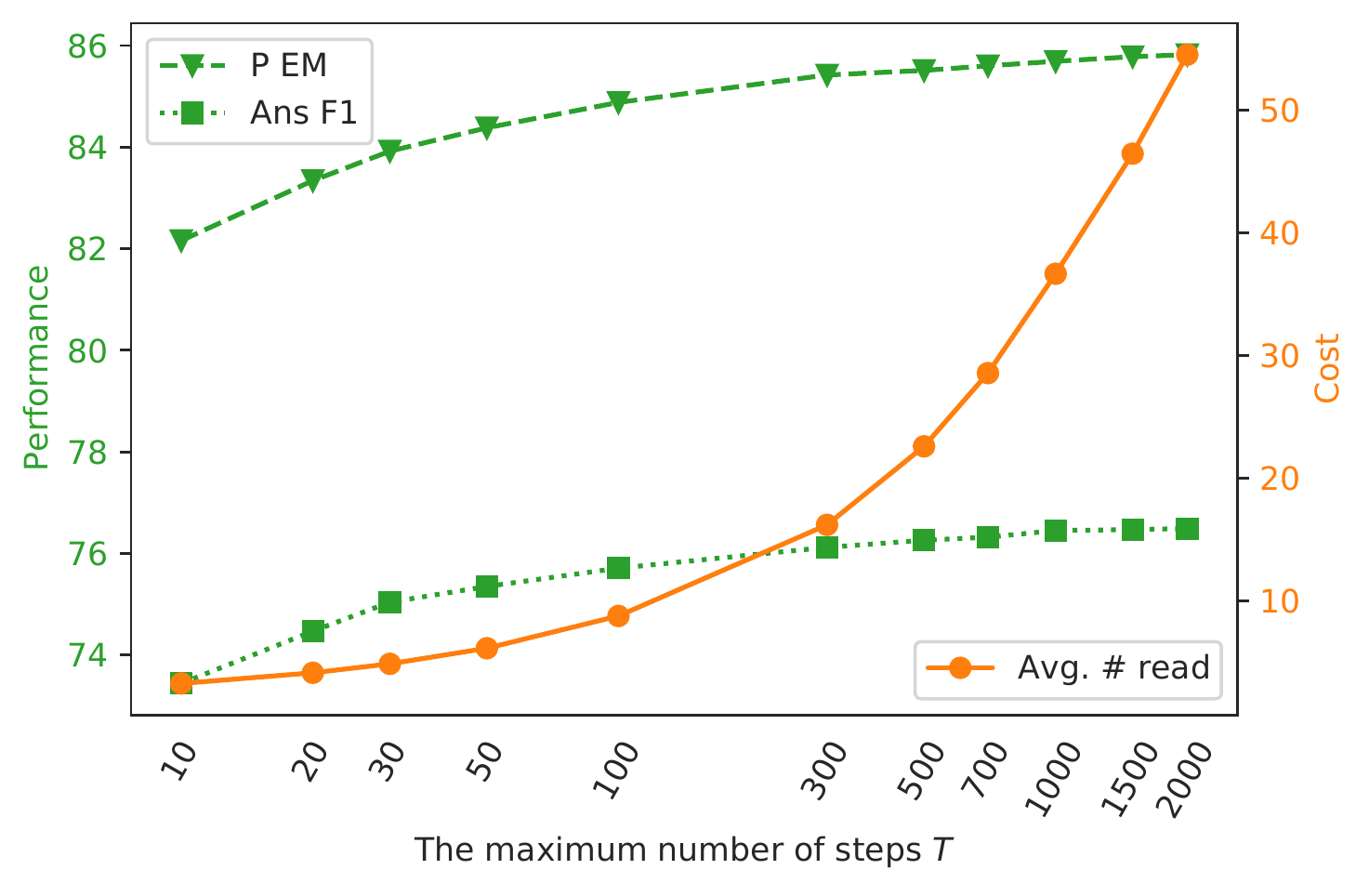}
    \caption{Performance and cost of AISO$_\mathrm{base}$ on the HotpotQA development set with different step limits.}
    \label{fig:step_metric}
\end{figure}

\paragraph{The impact of the maximum number of steps}
As shown in Figure~\ref{fig:step_metric}, with the relaxation of the step limit $T$, AISO$_\mathrm{base}$ can filter out negative passages and finally observe low-ranked evidence through more steps, so its performance improves and tends to converge.
However, the cost is more paragraphs to read.
% 500, 3\%
Besides, once $T$ exceeds 1000, only a few questions (about 1\%) can benefit from the subsequent steps.

\paragraph{The ability to recover from mistakes}
We count three types of mistakes in gathering evidence on the HotpotQA development set. 
In the process of collecting evidence for 7405 questions, false evidence was added into the evidence set for 1061 questions, true evidence was missed for 449 questions, and true evidence was deleted from the evidence set for 131 questions.
And we find that AISO recovered from 17.7\%, 43.9\%, and 35.9\% of these three types of errors respectively, which implies that even without beam search, AISO$_\mathrm{base}$ can make up for previous mistakes to some extent.
Besides, we can see that false evidence is the most harmful to evidence gathering and the most difficult to remedy.

\section{Conclusion and Future Work}
This work presents an adaptive information-seeking approach for open-domain question answering, called AISO. 
It models the open-domain QA task as a POMDP, where the environment contains a large corpus and the agent is asked to sequentially select retrieval function and reformulate query to collect the evidence. 
AISO achieves state-of-the-art results on two public datasets, which demonstrates the necessity of different retrieval functions for different questions. 
In the future, we will explore other adaptive retrieval strategies, like directly optimizing various information-seeking metrics by using reinforcement learning techniques.

\section*{Ethical Considerations}
We honor and support the ACL code of Ethics. The paper focuses on information seeking and question answering tasks, which aims to answer the question in the open-domain setting. It can be widely used in search engine and QA system, and can help people find the information more accuracy and efficiency. Simultaneously, the datasets we used in this paper are all from previously published works and do not involve privacy or ethical issues.

\section*{Acknowledgements}
This work was supported by National Natural Science Foundation of China (NSFC) under Grants No. 61906180, No. 61773362 and No. 91746301, National Key R\&D Program of China under Grants 2020AAA0105200.
The authors would like to thank Changying Hao for valuable suggestions on this work.
% and Wenhan Xiong for sharing the firsthand source code of MDR.
% We also thank the anonymous reviewers for their valuable suggestions.

\bibliography{anthology, custom}
\bibliographystyle{acl_natbib}

\end{document}